\documentclass[runningheads]{llncs}

\usepackage{eccv}

\usepackage{eccvabbrv}

\usepackage{graphicx}
\usepackage{booktabs}

\usepackage[accsupp]{axessibility}  
\usepackage{wrapfig}
\usepackage{multirow}
\usepackage{makecell}
\usepackage{adjustbox}
\usepackage{ulem}
\usepackage[misc]{ifsym}

\usepackage[pagebackref,breaklinks,colorlinks,citecolor=eccvblue]{hyperref}

\usepackage{orcidlink}

\newcommand{\myonedot}{\ifx\@let@token.\else.\null\fi\xspace}
\newcommand{\myetal}{\textit{et al}\myonedot}
\newcommand{\myeg}{\textit{e.g}\myonedot}
\newcommand{\myie}{\textit{i.e}\myonedot}

\begin{document}

\title{ET-SAM: Efficient Point Prompt Prediction in SAM for Unified Scene Text Detection and Layout Analysis} 

\titlerunning{ET-SAM}

\author{
Xike Zhang\inst{1}$^{\dagger}$\orcidlink{0009-0003-6364-8103} \and
Maoyuan Ye\inst{1}$^{\dagger}$\orcidlink{0000-0002-4180-1096} \and
Juhua Liu\inst{1}$^{(\textrm{\Letter})}$\orcidlink{0000-0002-3907-8820} \and
Bo Du\inst{1}\orcidlink{0000-0002-0059-8458}
}

\footnotetext{$\dagger$: Equal contribution. $\textrm{\Letter}$: Corresponding author.}

\authorrunning{X. Zhang et al.}

\institute{
School of Computer Science, National Engineering Research Center for Multimedia Software, Institute of Artificial Intelligence, and Hubei Key Laboratory of Multimedia and Network Communication Engineering, Wuhan University, China\\
\email{\{zhangxike, yemaoyuan, liujuhua, dubo\}@whu.edu.cn}
}

\maketitle

\begin{abstract}
Previous works based on Segment Anything Model (SAM) have achieved promising performance in unified scene text detection and layout analysis. 
However, the typical reliance on pixel-level text segmentation for sampling thousands of foreground points as prompts leads to \textit{unsatisfied inference latency} and \textit{limited data utilization}.
To address above issues, we propose \textbf{ET-SAM}, an \textbf{E}fficient framework with two decoders for unified scene \textbf{T}ext detection and layout analysis based on \textbf{SAM}. 
Technically, we customize a lightweight point decoder that produces word heatmaps for achieving a few foreground points, thereby eliminating excessive point prompts and accelerating inference. 
Without the dependence on pixel-level segmentation, we further design a joint training strategy to leverage existing data with heterogeneous text-level annotations. 
Specifically, the datasets with multi-level, word-level only, and line-level only annotations are combined in parallel as a unified training set. 
For these datasets, we introduce three corresponding sets of learnable task prompts in both the point decoder and hierarchical mask decoder to mitigate discrepancies across datasets.
Extensive experiments demonstrate that, compared to the previous SAM-based architecture, ET-SAM achieves about \textbf{3$\times$} inference acceleration while obtaining competitive performance on HierText, and improves an average of 11.0\% F-score on Total-Text, CTW1500, and ICDAR2015. The code and models are available at \href{https://github.com/zxk0228/ET-SAM}{ET-SAM}.
  \keywords{Segment Anything Model \and Unified Scene Text Detection \and Layout Analysis \and Point Prompt Prediction \and Joint Training}
\end{abstract}

\section{Introduction}
Visual text contained in images conveys rich and explicit semantic information which is pivotal for scene comprehension.
Meanwhile, visual text encompasses a hierarchical structure ranging from word and text-line to full paragraph, with each granular level serving a distinct role for varying applications~\cite{db++,abcnetv2,deepsolo,hi-sam,hiertext}.
However, previous scene text detection methods~\cite{textboxes++,db++,abcnetv2,textbpn++,dptext-detr,panet,ye2023deepsolo++, du2022i3cl} typically focus on a specific text level with an individual suit of parameters, lacking a unified framework for simultaneously tackling all text levels.
Consequently, the unified scene text detection and layout analysis benchmark~\cite{hiertext} has been introduced to holistically evaluate the performance across word, text-line, and paragraph levels. This benchmark is accompanied by a baseline method, the Unified Detector (UD), which leverages the Detection Transformer (DETR)~\cite{carion2020end} paradigm. 
Each object query in UD is optimized to segment individual words within a text line, while a dedicated layout branch constructs an affinity matrix across queries to facilitate their clustering into paragraphs. 

\begin{figure*}[t!]
    \centering
    \includegraphics[width=\linewidth]{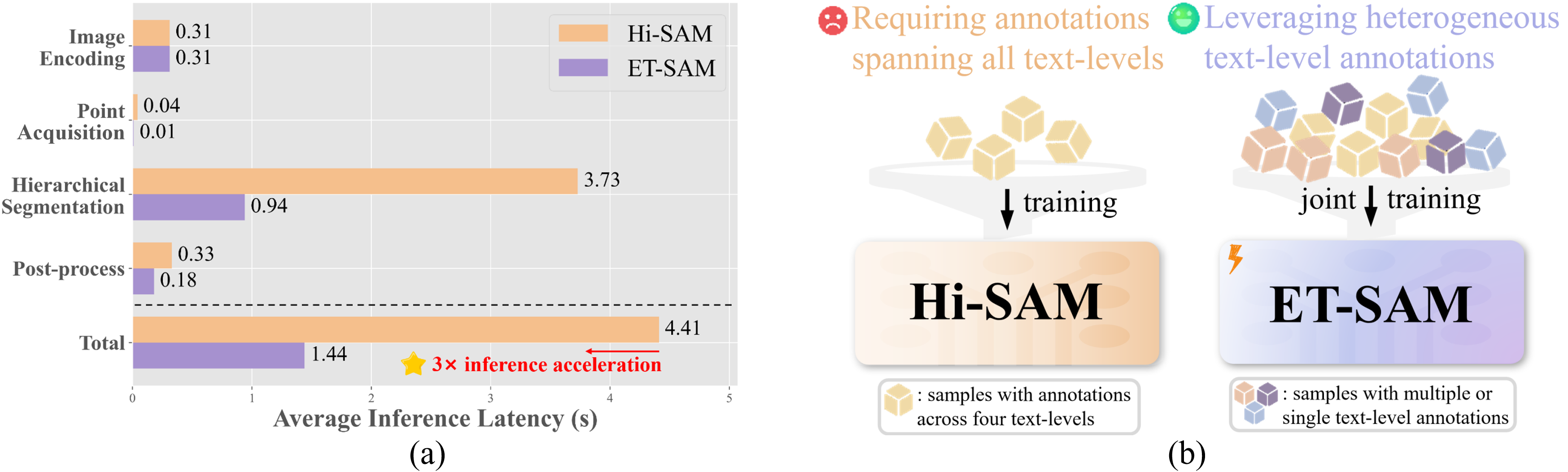}
    \caption{(a) ET-SAM achieves $3\times$ inference acceleration on HierText~\cite{hiertext} via predicting sparse points as visual prompts, significantly reducing the latency at hierarchical segmentation and post-processing stages. (b) To facilitate data scalability, we design a joint training strategy to leverage samples with mixing text-level annotations.}
    \label{fig:introduction}
    \vspace{-5mm}
\end{figure*}

Subsequently, built upon Segment Anything Model (SAM)~\cite{sam}, Hi-SAM~\cite{hi-sam} innovatively integrates four-level text segmentation and enhances the accuracy via a dual-decoder framework: one segments pixel-level text for generating foreground point prompts, while the other segments text structures from words to paragraphs under the guide of point prompts. 
However, Hi-SAM exhibits limitations in terms of \textit{inference latency} and \textit{data utilization}. Concretely, Hi-SAM randomly samples thousands of foreground points from the pixel-level segmentation result as visual prompts, introducing excessive inference cost at the hierarchical mask decoding and post-processing stages. Moreover, Hi-SAM necessitates comprehensive annotations spanning all text granularities for every training instance, a requirement that is constrained by the scarcity of such densely labeled data in current datasets. Thus, the scarcity of suitable training data impedes scalability, thereby constraining applications in real-world scenarios.

To address the issues, we present ET-SAM, a more efficient and scalable framework (as shown in Fig.~\ref{fig:introduction}) for unified scene text detection and layout analysis. 
Technically, to reduce the point prompts and speed up the inference, we introduce a customized lightweight point decoder that generates word-level heatmaps, enabling the efficient localization of sparse foreground points characterized by peak activation values. 
Moreover, we devise a joint training strategy that enables the combination of datasets with heterogeneous text-level annotations to facilitate data scalability. 
To disambiguate diverse word densities and segmentation targets while alleviating annotation inconsistencies, we further integrate distinct sets of learnable task prompts within the point decoder and mask decoder.

In summary, our main contributions are three-fold:
\begin{itemize}
    \item We present ET-SAM, enabling efficient point prompt prediction and speeding up the inference of SAM-based architecture with a tailored point decoder. 
    \item We devise a joint training strategy to harness existing datasets characterized by heterogeneous text-level annotations, thereby fostering the scaling of data.
    \item Extensive experiments demonstrate that ET-SAM not only delivers roughly \textbf{3$\times$} inference acceleration over Hi-SAM with competitive performance on HierText but also secures a substantial \textbf{11.0\%} gain in average F-score on Total-Text, CTW1500, and ICDAR2015.
\end{itemize}

\section{Related Work}

\subsection{Scene Text Detection}
Scene text detection has witnessed significant advancements in recent years, with most methodologies broadly classified into segmentation-based and regression-based paradigms.
Segmentation-based approaches~\cite{pse,panet,db,db++,srformer,ye2020textfusenet} formulate text detection as a dense pixel-wise segmentation task. 
For example, PANet~\cite{panet} introduces a pixel aggregation mechanism to associate pixels with distinct text kernels. DBNet~\cite{db} incorporates differentiable binarization to jointly predict threshold maps alongside binary segmentation masks. 
In comparison, regression-based methods~\cite{textboxes,textboxes++,east,abcnet,abcnetv2,fce,textbpn++,dptext-detr,deepsolo,he2024gomatching,he2026gomatching++} localize text regions via bounding box regression or contour curve fitting. 
For instance, ABCNet~\cite{abcnet} and DeepSolo~\cite{deepsolo} employ Bezier curves to model the top and bottom boundaries of text instances, whereas FCENet~\cite{fce} reconstructs contours using the inverse Fourier transform. 
Despite these strides, such methods typically operate at a singular text granularity with dedicated parameter sets, lacking a unified framework capable of addressing diverse text levels across words, text-lines, and paragraphs.

\subsection{Unified Scene Text Detection and Layout Analysis}
Long \myetal~\cite{hiertext} pioneered unified scene text detection and layout analysis by introducing the HierText dataset and the Unified Detector (UD). Built upon the DETR~\cite{carion2020end} paradigm, UD optimizes object queries for word and text-line segmentation while employing a layout branch to cluster them into paragraphs. Subsequently, Hi-SAM~\cite{hi-sam} advanced this field by integrating the Segment Anything Model~\cite{sam} for hierarchical text segmentation, enabling fine-grained segmentation from pixel-level strokes up to paragraphs. Nevertheless, Hi-SAM exhibits limitations regarding inference latency and data scalability. Specifically, its reliance on random point sampling incurs overheads, while requiring comprehensive multi-granularity annotations hinders scaling up training data. 
In comparison, we provide a more efficient and scalable framework. This work introduces a customized lightweight point decoder to predict sparse foreground points for acceleration, alongside a joint training strategy which leverages heterogeneous annotation sources. 
Note that the structure modifications are fully compatible to SAM components, which can also be applied to other potential SAM-based methods. 

\subsection{Segment Anything Model}
The Segment Anything Model (SAM)~\cite{sam} has established itself as a cornerstone of image segmentation, exhibiting exceptional generalization and adaptability across a spectrum of domains, ranging from image matting~\cite{li2024matting}, 3D segmentation~\cite{cen2023segment}, remote sensing~\cite{gao2025combining,SAMRS}, medical image segmentation~\cite{yue2023surgicalsam,yue2023part}, and video tracking~\cite{cheng2023segment,yang2023track}. 
Building upon this foundation, SAM 2~\cite{sam2} incorporates a streaming memory mechanism to enable unified segmentation and tracking for both static images and dynamic videos. 
Subsequently, SAM 3~\cite{sam3} further consolidates object detection, segmentation, and tracking into a single framework driven by concept prompts. 
Following the recent advance~\cite{hi-sam} in unified scene text detection and layout analysis, this work presents ET-SAM, a specialized and efficient architecture which is also based on SAM.

\section{Methodology}

\begin{figure*}[t]
    \centering
    \includegraphics[width=0.95\textwidth]{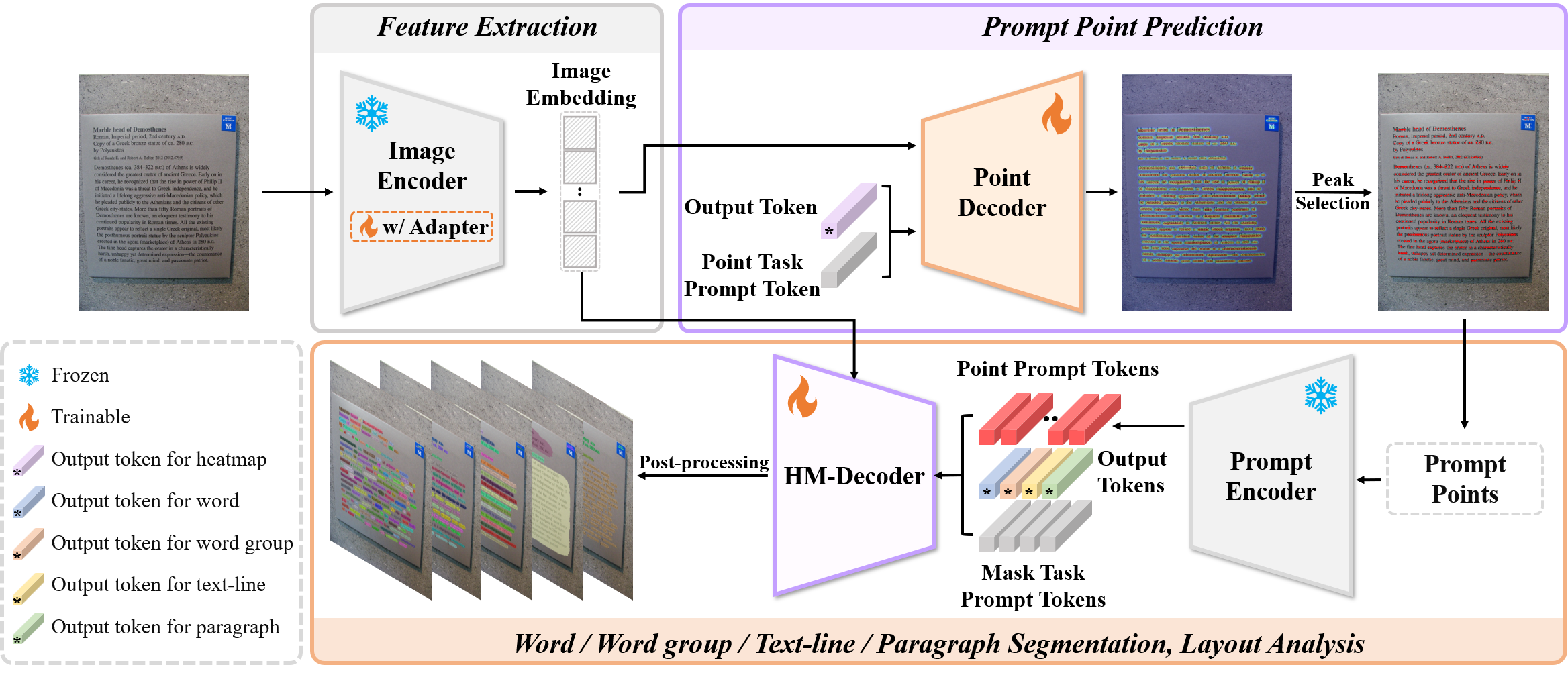}
    \caption{\textbf{Overview of ET-SAM framework.} 
    Extracted from the word heatmap produced by our point decoder, sparse points are used as visual prompts, thereby accelerating subsequent inference.
    Under the guide of point prompts and specific mask task prompts, HM-Decoder segments word, word group, text-line, and paragraph masks, which are finally used to achieve layout analysis via a union-find algorithm.}
    \label{fig:ET-SAM}
\end{figure*}

\subsection{Overview of ET-SAM}
The overall framework is depicted in \cref{fig:ET-SAM}.
ET-SAM comprises four primary components: (1) an image encoder adapted from SAM with adapters, aligned with the configuration of Hi-SAM~\cite{hi-sam}; (2) a streamlined, trainable point decoder derived from SAM's mask decoder, following PseCo \cite{pseco} paradigm; (3) a prompt encoder inherited directly from SAM; and (4) a hierarchical mask decoder (HM-Decoder), which is a specialized modification of SAM's mask decoder. 
Specifically, the point decoder generates a word-centric heatmap from image embeddings, where word centers exhibit peak activations. The coordinates of local maxima are then extracted as sparse points which are subsequently encoded by the prompt encoder and fed into the HM-Decoder to produce masks across hierarchical text granularities.
Leveraging these hierarchical masks, layout analysis is naturally derived as an auxiliary output, consistent with Hi-SAM paradigm. 
To orchestrate diverse segmentation tasks and alleviate inter-dataset discrepancies, both the point decoder and the HM-Decoder are augmented with learnable task prompts.
In \cref{sec:task prompt}, we first detail the task prompts within the point decoder (\cref{sec:point decoder}) and HM-Decoder (\cref{sec:h-decoder}), followed by descriptions of the joint training strategy (\cref{sec:joint training}) and inference workflow (\cref{sec:inference process}).

\begin{figure*}[t]
    \centering
    \includegraphics[width=0.8\textwidth]{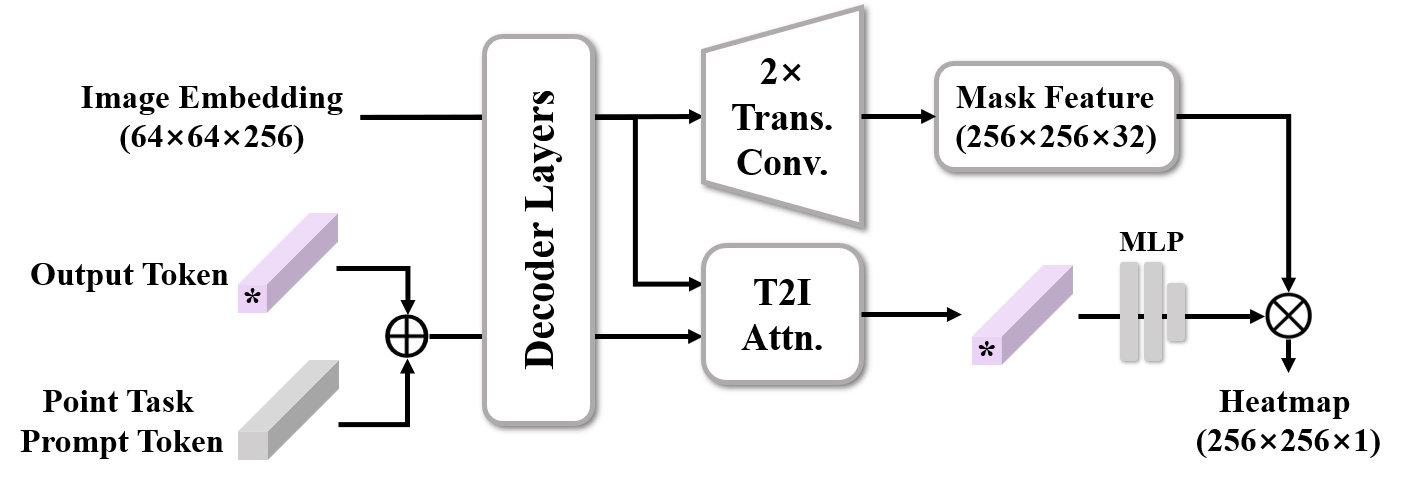}
    \caption{\textbf{Structure of Point Decoder.} `Trans. Conv.' and `T2I Attn.' denote transposed convolution and token-to-image attention, respectively.
    }
    \label{fig:point_decoder}
\end{figure*}

\subsection{Task Prompt}\label{sec:task prompt}
Heterogeneity across datasets poses a challenge to model optimization. For instance, HierText~\cite{hiertext} exhibits high word density and employs quadrilateral annotations for text lines, whereas CTW1500~\cite{ctw1500} utilizes curved representations for text lines. Such disparities may degrade cross-dataset performance.
To mitigate these discrepancies, in both the point decoder and the HM-Decoder, we incorporate learnable task prompts that are initialized with simple embeddings.
These prompts serve as conditional signals: in the point decoder, they adapt to varying word densities, while in the HM-Decoder, they specify distinct segmentation objectives.
Specifically, we define three task configurations ($N_{task}=3$):
\begin{itemize}
    \item \textit{Task 0} targets multi-level datasets (\myeg, HierText), featured with dense word distributions and requiring simultaneous prediction of hierarchical masks;
    \item \textit{Task 1} necessitates only word-level mask prediction;
    \item \textit{Task 2} focuses on text-line level mask prediction.
\end{itemize}

\subsection{Point Decoder}\label{sec:point decoder}
The proposed point decoder is a streamlined variant derived from SAM's original mask decoder.
Specifically, we architecturally refine the module by: (1) replacing visual prompts with learnable point task prompt tokens; (2) consolidating the multiple Multi-Layer Perceptrons (MLPs) designed for multi-granularity mask generation into a singular MLP; and (3) removing the Intersection-over-Union (IoU) prediction head. Consequently, the decoder directly outputs a single word-centric heatmap from image embeddings. 
The overall architecture is depicted in \cref{fig:point_decoder}, where the Transformer blocks and transposed convolution layers retain the pre-trained weights from SAM's mask decoder.

\begin{wrapfigure}{r}{0.4\textwidth}
    \centering
    \vspace{-5mm}
    \includegraphics[width=0.40\textwidth]{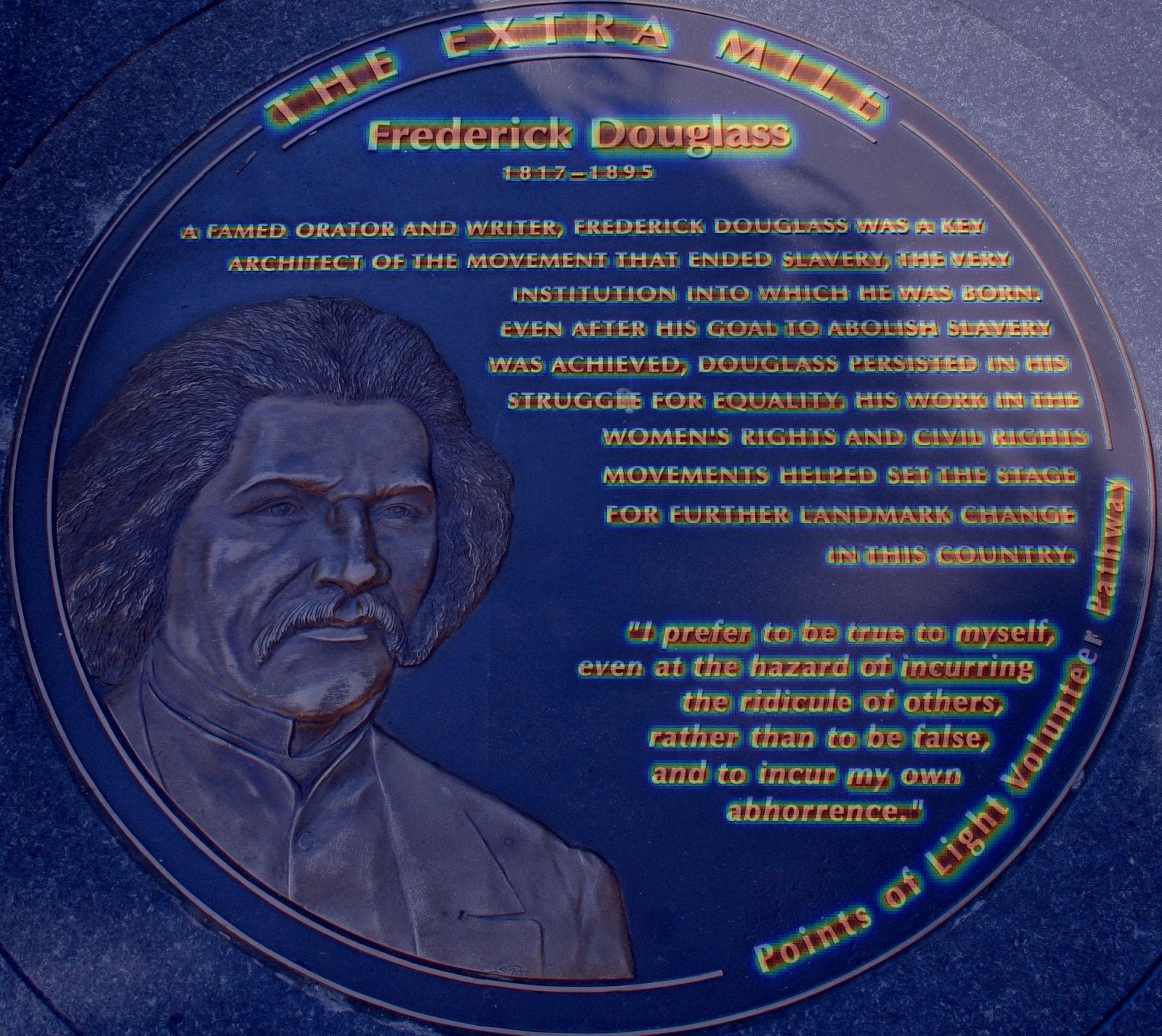}
    \caption{\textbf{A visualization of target word-centric Heatmap.}}
    \vspace{-5mm}
    \label{fig:target_heatmap}
\end{wrapfigure}

Formally, let $t_{p\_out} \in \mathbb{R}^{1 \times 256}$ denote the learnable output token and $t_{p\_task} \in \mathbb{R}^{N_{task} \times 256}$ represent the set of learnable point task prompt tokens, with $N_{task}$ indicating the total number of tasks.
For a selected task $i$, the output token $t_{p\_out}$ is modulated by the task prompt token $t_{p\_task}[i]$ via element-wise addition, yielding an aggregated token $t_{p\_out}'$ fed into the point decoder.
Following the token-to-image attention and transposed convolution operations, we obtain the refined output token $\hat{t}_{p\_out} \in \mathbb{R}^{1 \times 256}$ and the intermediate mask feature map $F_p \in \mathbb{R}^{256 \times 256 \times 32}$.
Next, $\hat{t}_{p\_out}$ is projected through an MLP to align its dimensionality with $F_p$.
Finally, the output of the MLP is linearly projected onto $F_p$ to generate the heatmap $\hat{H} \in [0,1]^{256 \times 256}$.

\noindent\textbf{Word Heatmap Label Generation.} We construct the target heatmap $H \in [0,1]^{256 \times 256}$ derived from ground-truth word annotations.
Specifically, the word center lines are first determined from the annotations and dense points are sampled along these center lines.
For each sampled point $P(P_x,P_y)$, a local Gaussian kernel is generated as follows: 

\begin{equation}
H_{xy} = \exp\left(
- \frac{(x - p_x)^2 + (y - p_y)^2}{2\sigma_w^2}
\right),
\end{equation}
where $(p_x,p_y)=(\frac{P_x}{4},\frac{P_y}{4})$ denote the downsampled coordinates of point $P$ on the heatmap grid, $(x,y)$ represents an arbitrary spatial coordinate on the heatmap, $\sigma_w$ is the standard deviation adjusted according to the width of the word contour.
Finally, the heatmaps of all sampled points across all words are merged to produce the target heatmap $H$, taking the maximum in each grid location as the final value on $H$. An example of target heatmap is shown in \cref{fig:target_heatmap}.

\subsection{HM-Decoder}\label{sec:h-decoder}
\begin{figure*}[t!]
    \centering
    \includegraphics[width=0.95\textwidth]{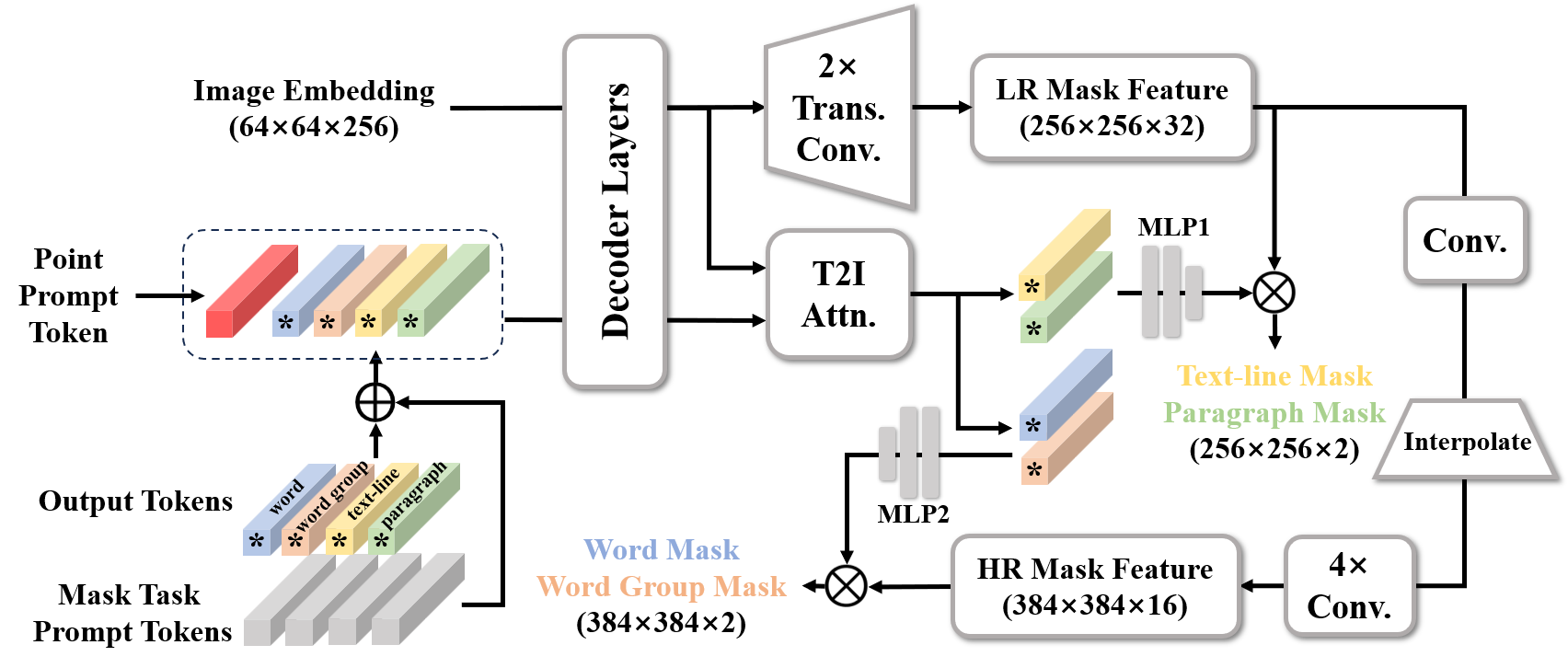}
    \caption{\textbf{Structure of HM-Decoder.} 
    Four output tokens charge in segmentation at word, word group, text-line, and paragraph level, under the guide of point prompts and learnable mask task prompt tokens.
    `Interpolate' indicates interpolating the spatial resolution to $384\times384$. The IoU prediction process is omitted here.}
    \label{fig:h-decoder}
    \vspace{-3mm}
\end{figure*}

For each point prompt, the HM-Decoder is tailored to simultaneously segment four text granularities, \myie, an individual word, intra-line words, a complete text-line, and a paragraph (along with predicting their respective IoU scores).
The detailed structure is illustrated in \cref{fig:h-decoder}.

Concretely, the output tokens $t_{h\_out} \in \mathbb{R}^{1 \times 4 \times 256}$ and the mask task prompt tokens $t_{h\_task} \in \mathbb{R}^{N_{task} \times 4 \times 256}$ are learnable embeddings, where $N_{task}$ denotes the number of tasks while the dimension of 4 corresponds to the number of output masks.
Given $K$ point prompt tokens $t_{point} \in \mathbb{R}^{K \times N_{p} \times 256}$ encoded by the prompt encoder and a selected task $i$, we first compute the element-wise sum of $t_{h\_out}$ and the $i$-th task prompt token $t_{h\_task}[i]$. The resulting tensor is subsequently broadcast to align with the number of point prompt tokens $t_{point}$, yielding an intermediate representation $t_{h\_out}' \in \mathbb{R}^{K \times 4 \times256}$.
Then, $t_{h\_out}'$ is concatenated with $t_{point}$, forming $[t_{h\_out}'; t_{point}] \in \mathbb{R}^{K \times (4+N_p) \times 256}$ which is fed into the HM-Decoder.

After the transposed convolution and token-to-image attention operations, the updated output tokens $\hat{t}_{h\_out} \in \mathbb{R}^{K \times 4 \times 256}$ and low-resolution mask features $F_{lr} \in \mathbb{R}^{K\times256\times256\times32}$ are produced.
To generate the mask logits $M_{l,p}\in \mathbb{R}^{K\times256\times256\times2}$ for text-line and paragraph hierarchies, the corresponding slice of output tokens $\hat{t}_{h\_out}$ is projected via a MLP subsequently fused with the low-resolution mask features $F_{lr}$ through a spatially point-wise product. 
Conversely, for the word and word group hierarchies, the corresponding sliced output tokens are processed by another MLP.
These transformed output tokens are then combined with high-resolution features $F_{hr} \in \mathbb{R}^{K \times 384 \times 384 \times 16}$, which are derived from $F_{lr}$ via interpolation and convolutional refinement, to produce the final word and word group mask logits $M_{w,wg}\in \mathbb{R}^{K \times 384 \times 384 \times 2}$.
Consistent with the design of Hi-SAM~\cite{hi-sam}, we generate word and word group masks at an elevated spatial resolution for better segmentation quality.

\subsection{Joint Training}\label{sec:joint training}
Acknowledging the scarcity of multi-level annotations beyond HierText~\cite{hiertext}, we stratify existing datasets into word-level, text-line-level, and multi-level categories. Our objective is to jointly leverage these datasets to achieve uniform representation learning across text hierarchies.
However, naive joint training poses challenges: if individual batches are dominated by samples from a single dataset or text granularity, the optimization trajectory may fluctuate significantly between batches.
To mitigate these issues, we first preprocess datasets within each category into a unified format, then randomly shuffle and aggregate them to construct three consolidated datasets: a multi-level set, a word-level set, and a text-line-level set.
These sets are subsequently aligned in parallel to form a unified training pool, with its total length dictated by the largest set.
During joint training, each batch is constructed to include one sample from each of the three categories.
Within every epoch, the shorter datasets are iteratively resampled until they match the length of the unified dataset.

The trainable components of our framework comprise: (1) the adapters integrated within the image encoder, (2) the point decoder, and (3) the HM-Decoder.
Specifically, the optimization of the point decoder is governed by the  Mean Squared Error (MSE) loss, which quantifies the discrepancy between the predicted heatmaps $\hat{H}$ and the ground-truth targets $H$. This loss function is formulated as: $\mathcal{L}_{point} =  \lVert H - \hat{H} \rVert_2^2$.

For optimizing HM-Decoder, a certain number of prompt points and their corresponding masks are first sampled.
For HierText dataset, we randomly sample 10 text-lines from each image, obtain their corresponding word groups and paragraphs, and then randomly select one word from each word group. 
Next, we extract extreme points from the target heatmap and take their intersections with the word masks. 
Similarly, for word- and text-line–level datasets, we randomly sample 10 words or text-lines. 
We then extract extreme points from the heatmaps and take their intersections with the corresponding masks. 
Finally, for each mask, up to 2 points are randomly selected as prompt points, yielding at most 20 prompt points paired with ground-truth masks across four text granularities.
These prompt points are fed into the prompt encoder to obtain point prompt tokens, guiding the segmentation of hierarchical masks.
The losses for word, word group, text-line, and paragraph masks are all composed of Binary Cross-Entropy (BCE) loss, Dice loss~\cite{dice}, and IoU MSE loss, with an equal weighting ratio of 1:1:1.
The overall loss function is defined as follows:
\begin{equation}
    \mathcal{L} = 50 \times \mathcal{L}_{point} + \mathcal{L}_{word} + \mathcal{L}_{word\_group} + \mathcal{L}_{line} + 0.5 \times \mathcal{L}_{para}.
\end{equation}

Additionally, text-line datasets do not contain word-level ground truth, which conflicts with the word-centric nature of our heatmap generation process described in \cref{sec:point decoder}.
To resolve this discrepancy, we first train the point decoder using multi-level and word-level datasets, then use the trained model to generate word heatmaps as pseudo labels. 
These heatmaps are further refined using available text-line annotations to eliminate errors, such as suppressing false positives.

\subsection{Inference Process}\label{sec:inference process}
For a given image and selected task, the point decoder initially generates a word-level heatmap.
A $3 \times 3$ max-pooling operation is then applied to extract extrema from the heatmap, and points whose values exceed a predefined threshold are selected as prompt points. 
These points are fed into the prompt encoder in batches of 100 to obtain point prompt tokens, which guide the HM-Decoder to segment masks and predict IoU scores across four hierarchical levels.

For \textit{task 0}, we first filter out text-lines with low IoU and apply Matrix NMS~\cite{nms} to suppress overlapping text-lines. 
The corresponding word groups and paragraphs associated with the removed text-lines are also discarded. 
The IoU and NMS threshold are both set to 0.5 by default.
Then, we compute the IoU matrix of the paragraph masks and use a union–find algorithm to realize layout analysis following previous works~\cite{hiertext,hi-sam}. 
Finally, the corresponding word groups and text lines are aggregated accordingly, completing the layout analysis~\cite{hiertext}.

For \textit{tasks 1 and 2}, we first discard low-quality masks whose IoU scores are below 0.5, and then apply Matrix NMS with a threshold of 0.5 to eliminate overlapping masks, obtaining the final word or text-line masks.

\section{Experiments}
\subsection{Datasets}
\textbf{HierText}~\cite{hiertext} comprises 8,281 training, 1,724 validation, and 1,634 test images. 
It is distinguished by its hierarchical annotations spanning word, text-line, and paragraph levels.
Statistically, each image contains hundreds of words, exhibiting a distribution that is both sufficiently dense and relatively uniform.

\noindent \textbf{Total-Text}~\cite{total} contains 1,255 training images and 300 test images, featuring a large number of curved text instances and providing word-level annotations.

\noindent \textbf{TextSeg}~\cite{textseg} encompasses 2,646 training, 340 validation, and 1,038 test images. TextSeg offers both word-level and pixel-level annotations, covering a diverse array of scene and artistic text instances.

\noindent \textbf{ICDAR2013}~\cite{ic13} contains 229 training and 233 test images, focusing primarily on horizontal scene text with word-level bounding box annotations. 
Compared to ICDAR2013, \textbf{ICDAR2015}~\cite{ic15} presents a more challenging scenario with 1,000 training and 500 test images. ICDAR2015 incorporates multi-oriented text instances, which are annotated with quadrilateral boxes.

\noindent \textbf{CTW1500}~\cite{ctw1500} consists of 1,000 training and 500 test images, featuring the text-line level benchmark with annotations in arbitrary shapes.

\subsection{Implementation Details}
\textbf{Joint Training.} The majority parameters of ET-SAM are initialized from SAM before training. 
Specifically, the image encoder adopts SAM’s ViT-L~\cite{sam}. The prompt encoder is identical to that of SAM, while the point decoder and HM-Decoder are partially initialized from SAM's mask decoder.
We use AdamW~\cite{adamw} $(\beta_1 =0.9,\beta_2 =0.999,weight\_decay =0.05)$ as the optimizer with an initial learning rate of $1e^{-4}$.
We train the model on 8 NVIDIA Tesla V100 (32GB) GPUs, with a batch size of 3 per GPU.
We follow the data augmentations used in Hi-SAM~\cite{hi-sam}. 
Joint training runs for 120 epochs on a unified data pool including HierText, Total-Text, TextSeg, ICDAR2013, ICDAR2015, and CTW1500, with the learning rate divided by 10 at the 100 epochs.

\noindent \textbf{Fine-tuning.} We fine-tune the model after joint training on 4 NVIDIA Tesla V100 (32GB) GPUs with a total batch size of 8, using the same optimizer and data augmentation strategy, and an initial learning rate of $1e^{-4}$.
On HierText, ET-SAM is trained for 50 epochs, with the learning rate decayed by a factor of 10 at the 35th epoch. 
On Total-Text, CTW1500, and ICDAR2015, it is trained for 80 epochs, with the learning rate divided by 10 after 60 epochs.

\begin{table*}[t!]
\centering
\caption{\textbf{Performance comparison on the test set of HierText.}}
\label{tab:main results on HierText test}
\begin{adjustbox}{max width=\textwidth}
\begin{tabular}{l|ccccc|ccccc|ccccc}
\toprule
\multirow{2}{*}{Method} 
& \multicolumn{5}{c|}{Word}
& \multicolumn{5}{c|}{Text-line}
& \multicolumn{5}{c}{Layout Analysis} \\

\cmidrule(lr){2-6} \cmidrule(lr){7-11} \cmidrule(lr){12-16}
& PQ & F & P & R & T
& PQ & F & P & R & T
& PQ & F & P & R & T \\
\midrule

UD \cite{hiertext}
& 48.21 & 61.51 & 67.54 & 56.47 & \textbf{78.38}
& 62.23 & 79.91 & 79.64 & \uline{80.19} & 77.87 
& 53.60 & 68.58 & 76.04 & 62.45 & \textbf{78.17} \\

Hi-SAM-L \cite{hi-sam}
& \uline{63.10} & \uline{81.83} & \textbf{87.22} & \uline{77.06} & 77.11 
& \uline{66.17} & \textbf{84.85} & \textbf{90.66} & 79.74 & \uline{77.99} 
& 57.61 & 74.49 & 80.43 & \uline{69.37} & 77.34 \\
\midrule

Ours (Joint) 
& 62.52 & 80.84 & 85.01 & 77.06 & 77.33 
& 65.40 & 84.08 & 88.90 & 79.75 & 77.79 
& \uline{57.74} & \uline{74.51} & \uline{80.75} & 69.16 & 77.49 \\

Ours (FT) 
&  \textbf{63.60}&  \textbf{81.94}&  \uline{85.81}&  \textbf{78.41}&  \uline{77.61}
&  \textbf{66.30}&  \uline{84.82}&  \uline{89.17}&  \textbf{80.88}&  \textbf{78.16}
&  \textbf{58.82}&  \textbf{75.39}&  \textbf{81.85}&  \textbf{69.88}&  \uline{78.02}\\

\bottomrule
\end{tabular}
\end{adjustbox}
\vspace{-4mm}
\end{table*}

\subsection{Experimental Results}\label{sec:experimental results}
The main experimental results are evaluated on the HierText~\cite{hiertext}, Total-Text~\cite{total}, CTW1500~\cite{ctw1500}, and ICDAR2015~\cite{ic15} test sets, considering both the joint training results and the fine-tuning results.
For HierText, we follow previous works to use Panoptic Quality (PQ)~\cite{pq}, F-score (F), Precision (P), Recall (R), and Tightness (T) as metrics, while F-score, Precision, and Recall are used for other test sets.

\noindent\textbf{Unified Scene Text Detection and Layout Analysis.} 
Table \ref{tab:main results on HierText test} compares the comprehensive performance on the test set of HierText. Under the joint training protocol, ET-SAM exhibits a marginal deficit compared to Hi-SAM~\cite{hi-sam} at the word and text-line levels, yet surpasses it in layout analysis. After fine-tuning, ET-SAM achieves consistent gains across all hierarchies, with improvements of 1.08\% PQ and 1.10\% F-score at the word level, 0.90\% PQ and 0.74\% F-score at the text-line level, and 1.08\% PQ and 0.88\% F-score on layout analysis, resulting in overall superior performance compared to Hi-SAM.

\begin{table*}[t!]
\centering
\caption{\textbf{Performance comparison on Total-Text, CTW1500, and ICDAR15.}}
\label{tab:main results on TCI test}

\setlength{\tabcolsep}{6pt}
\resizebox{\linewidth}{!}{\begin{tabular}{l|ccc|ccc|ccc}
\toprule
\multirow{2}{*}{Method}
 & \multicolumn{3}{c|}{Total-Text} 
 & \multicolumn{3}{c|}{CTW1500} 
 & \multicolumn{3}{c}{ICDAR15} \\

\cmidrule(lr){2-4} \cmidrule(lr){5-7} \cmidrule(lr){8-10}
& F & P & R
& F & P & R
& F & P & R \\

\midrule
PANet \cite{panet}
&85.0	&89.3	&81.0	&83.7	&86.4	&81.2	&82.9	&84.0	&81.9 \\

DBNet++ \cite{db++}
&86.0  &88.9 &83.2 &85.3 &87.9 &82.8 &87.3 &90.9 &83.9 \\

PCR \cite{pcr}
&85.2	&88.5	&82.0	&84.7	&87.2	&82.3	&-	&-	&- \\

ABCNet v2 \cite{abcnetv2}
&87.0	&90.2	&84.1	&84.7	&85.6	&83.8	&88.1	&90.4	&86.0\\

FCENet \cite{fce}
&85.8&89.3&82.5&85.5&87.6&83.4&86.2&90.1&82.6\\

TESTR \cite{testr}
&86.9	&\textbf{93.4}	&81.4	&87.1	&\uline{92.0}	&82.6	&90.0	&90.3	&\textbf{89.7}\\

TextBPN++ \cite{textbpn++}
&88.5	&91.8	&85.3	&85.5	&87.3	&83.8	&-	&-	&-\\

DPText-DETR \cite{dptext-detr}
&89.0	&91.8	&86.4	&88.8	&91.7	&86.2	&-	&-	&-\\

DeepSolo \cite{deepsolo}
&\textbf{90.0}	&\uline{92.9}	&87.4	&88.9	&\textbf{93.2}	&85.0	&\uline{90.1}	&\uline{92.4}	&87.9\\

SRFormer \cite{srformer}
&89.7	&91.5	&87.9	&\uline{89.6}	&89.4	&\textbf{89.8}	&-	&-	&- \\

ESTextSpotter \cite{estextspotter}
&\uline{90.0}&92.0&\uline{88.1}&\textbf{90.0}&91.5&\uline{88.6}&\textbf{91.0}&\textbf{92.5}&\uline{89.6}\\

DocSAM \cite{docsam}
&77.0&72.1&82.6&84.2&80.5&88.1&-&-&-\\

UD \cite{hiertext}
& 87.9 & 85.0 & \textbf{91.1} & 86.0 & 84.6 & 87.4 & - & - & - \\

Hi-SAM-L (zero-shot)  \cite{hi-sam}
& 70.2 & 65.5 & 75.6& 77.4 & 80.1 & 74.9& 77.4 & 70.3 & 86.1 \\

\midrule
Ours (Joint) 
& 86.4 & 86.0 & 86.7& 87.1& 88.1& 86.2&  83.5&  85.0& 82.1\\

Ours (FT) 
&87.0 	&86.4 	&87.5 	&87.9&87.3&88.4&83.2&80.6&85.9\\

\bottomrule
\end{tabular}}
\end{table*}

\begin{figure*}[t!]
    \centering
    \includegraphics[width=\textwidth]{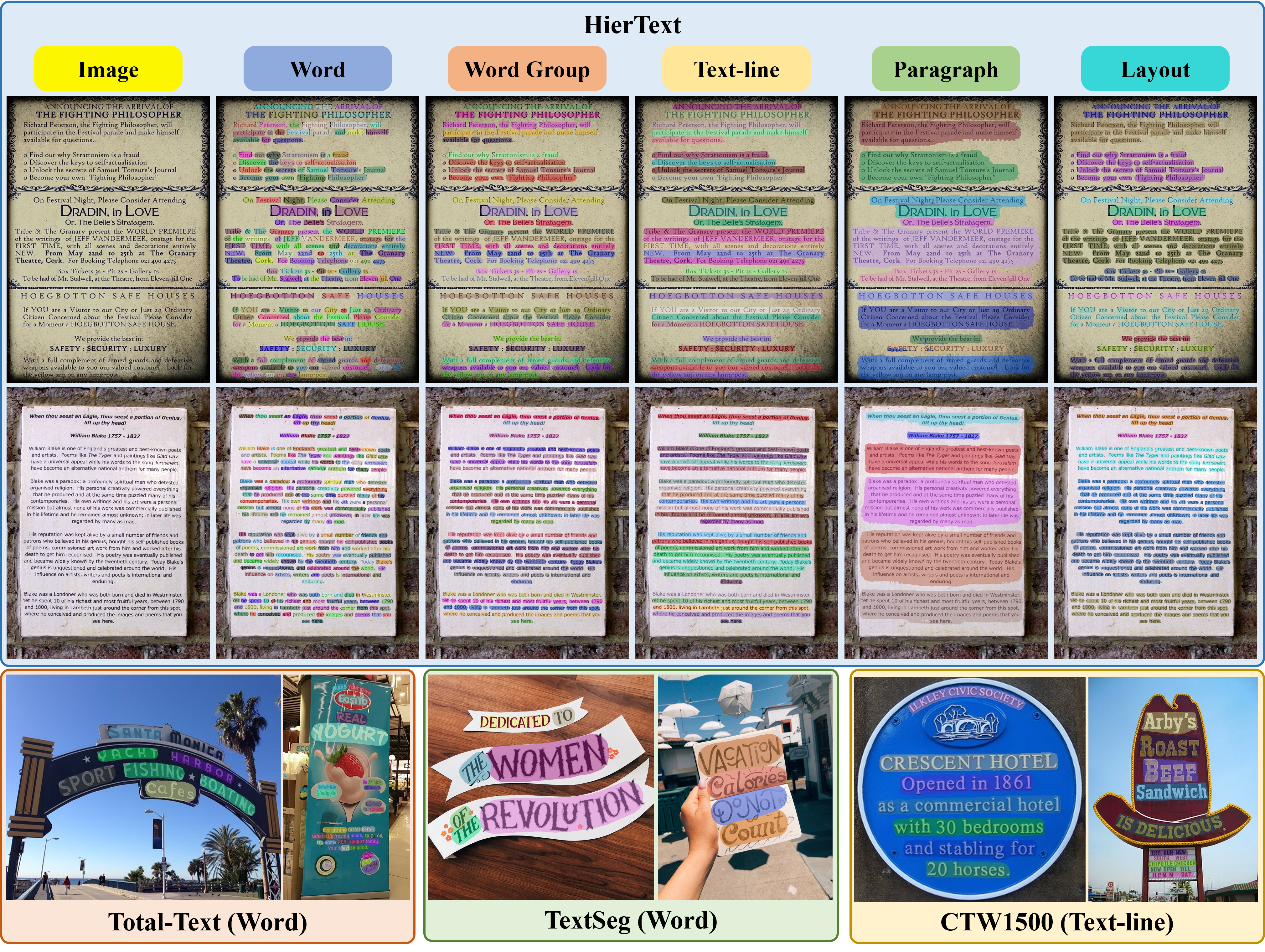}
    \caption{\textbf{Visualizations of scene text detection and layout analysis.} }
    \label{fig:visualize}
    \vspace{-4mm}
\end{figure*}

\noindent\textbf{Scene Text Detection on Specific Text Granularities.} 
As shown in \cref{tab:main results on TCI test}, we evaluate the text detection performance of different text granularities on Total-Text, CTW1500, and ICDAR2015 against state-of-the-art (SOTA) methods.
In contrast to Hi-SAM, which cannot leverage mixed-granularity annotations during training, our proposed ET-SAM demonstrates substantial improvements across all three benchmarks following the joint training stage, yielding an average F-score gain of 10.7\%.
Upon dataset-specific fine-tuning, ET-SAM achieves F-scores of 87.0\% on Total-Text, 87.9\% on CTW1500, and 83.2\% on ICDAR2015, culminating in a remarkable 11.0\% average improvement over Hi-SAM.
However, a performance disparity persists between our unified framework and specialized SOTA detectors such as DPText-DETR and SRFormer. 
We attribute this gap primarily to discrepancies in training data. In the appendix, we demonstrate that ET-SAM can finally surpass DPText-DETR by an average of 1.2\% F-score across benchmarks with the same training data. In the future, we could further scale up the training data volume for ET-SAM to obtain superior performance.

\begin{figure*}[t!]
    \centering
    \includegraphics[width=0.8\textwidth]{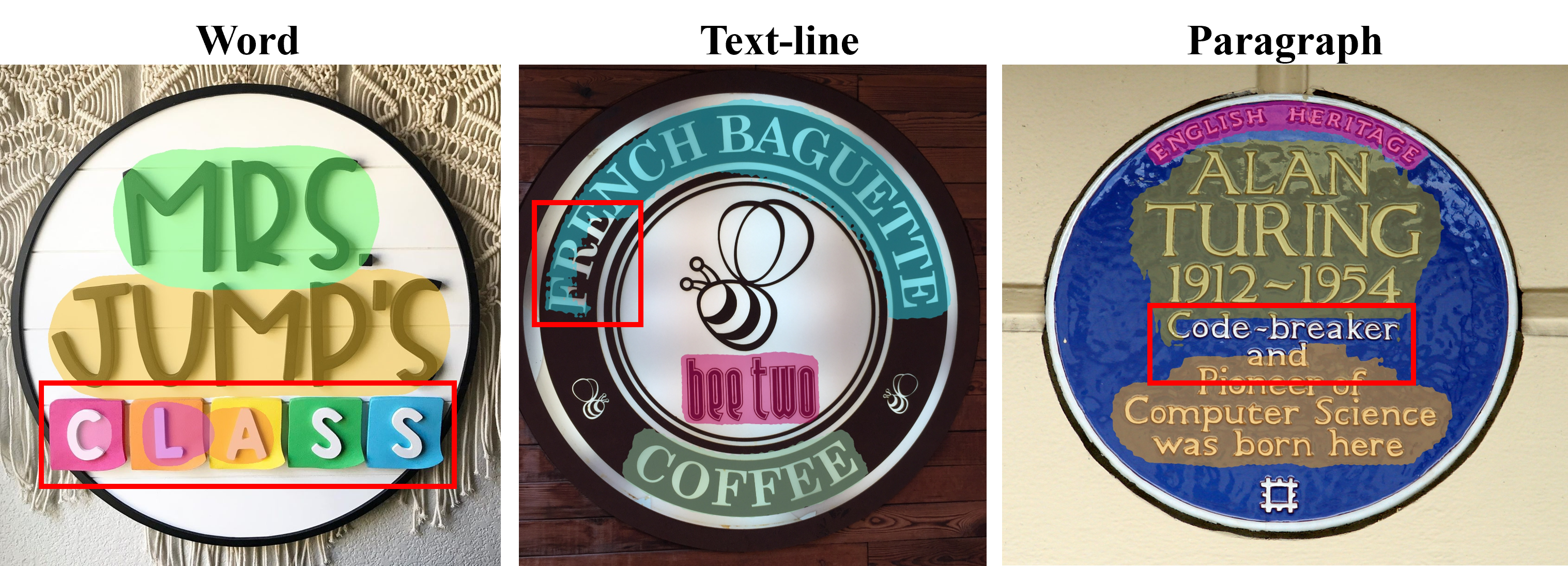}
    \caption{\textbf{Visualizations of failure cases.} }
    \label{fig:visualize_failure}
\end{figure*}

\noindent\textbf{Visualizations.} 
We provide visualizations across different datasets in \cref{fig:visualize} and \cref{fig:visualize_failure}, including several failure cases. Overall, most word- and text-line-level masks are accurate and of high quality. However, some word instances are partially incomplete when the background is visually inconsistent, and imperfections remain in text-line and paragraph-level segmentation results.

\begin{table*}[t]
\centering
\caption{\textbf{Effectiveness of the point decoder.} `$\dagger$' indicates replacing the S-Decoder in Hi-SAM with our designed point decoder for producing foreground points.
}
\label{tab:point decoder}

\setlength{\tabcolsep}{6pt}
\resizebox{\linewidth}{!}{\begin{tabular}{l|cc|cc|cc|c|c}
\toprule
\multirow{2}{*}{Method}
 & \multicolumn{2}{c|}{Word} 
 & \multicolumn{2}{c|}{Text-line} 
 & \multicolumn{2}{c|}{Layout} 
 & \multirow{2}{*}{\makecell{Average\\Points}}
 & \multirow{2}{*}{\makecell{Average\\Time}}\\

\cmidrule(lr){2-3} \cmidrule(lr){4-5} \cmidrule(lr){6-7}
& PQ & F & PQ & F & PQ & F & & \\

\midrule
Hi-SAM-L 
&  63.10&  \uline{81.83}&  66.17&  \uline{84.85}&  57.61&  74.49&  1500&  4.4s\\

Hi-SAM-L
&   61.04&  78.82&  62.29&   79.30&  54.59&  70.43&  280&  \uline{1.2s}\\

Hi-SAM-L$^{\dagger}$
& \uline{63.50}& 81.74& \textbf{66.96}& \textbf{85.30}& \uline{58.35}& \uline{74.63}&  \textbf{268}&  \textbf{1.2s}\\

Ours (Joint)
& 62.52& 80.84& 65.40& 84.08& 57.74& 74.51&  287&  1.4s\\

Ours (FT)
&\textbf{63.60}&\textbf{81.94}&\uline{66.30}&84.82&\textbf{58.82}&\textbf{75.39}&\uline{286}& 1.4s\\

\bottomrule
\end{tabular}}
\vspace{-4mm}
\end{table*}

\subsection{Ablation Studies}
\textbf{Point Decoder.} 
In \cref{tab:point decoder}, we validate the effectiveness of the point decoder.
Hi-SAM randomly samples 1500 foreground points from pixel-level text segmentation outputs of its S-Decoder. While reducing the sampling quota improves efficiency, it results in significant performance degradation.
In contrast, substituting the S-Decoder with the point decoder yields the variant Hi-SAM-L$^{\dagger}$, which achieves superior performance alongside a 3.6$\times$ inference acceleration (reducing latency from 4.4s to 1.2s). In terms of performance, Hi-SAM-L$^{\dagger}$ consistently outperforms Hi-SAM-L, especially achieving higher PQ and F-score at the text-line and layout hierarchies, indicating more accurate and informative visual prompts.
By further customizing the hierarchical mask decoder and applying joint training, our full method maintains competitive performance and high efficiency.
Overall, predicting sparse points by the point decoder as visual prompts serves as a robust and efficient alternative to Hi-SAM's random sampling, facilitating significant computational savings without compromising the performance.

\begin{table*}[t]
\centering
\caption{\textbf{Ablations on task prompts used in point decoder and HM-Decoder.}}
\label{tab:task prompt}

\setlength{\tabcolsep}{2pt}
\begin{adjustbox}{max width=\textwidth}
\begin{tabular}{cc|cccccc|ccc|ccc}

\toprule
\multirow{3}{*}{Point}
& \multirow{3}{*}{Mask}
& \multicolumn{6}{c|}{HierText}
& \multicolumn{3}{c|}{Total-Text}
& \multicolumn{3}{c}{CTW1500} \\
\cmidrule(lr){3-8} \cmidrule(lr){9-11} \cmidrule(lr){12-14}
&
& \multicolumn{2}{c}{Word}
& \multicolumn{2}{c}{Text-line}
& \multicolumn{2}{c|}{Layout}  
& \multicolumn{3}{c|}{Word}
& \multicolumn{3}{c}{Text-line} \\
\cmidrule(lr){3-4} \cmidrule(lr){5-6} \cmidrule(lr){7-8} \cmidrule(lr){9-11} \cmidrule(lr){12-14}
&
& PQ & F & PQ & F & PQ & F  
& F & P & R & F & P & R  \\
\midrule
 $\times$& $\times$
 & 62.35& \uline{80.87}& 65.08& 83.69& 56.77&73.38
 & 81.99& 78.54& 85.77& \uline{86.79}& 85.76&\textbf{87.83}\\
 
 $\times$& $\checkmark$
 & 62.36& 80.79& \uline{65.45}& 83.96& 57.23&73.76
 & 84.13& 81.85& \uline{86.54}& 86.30& 85.12&\uline{87.50}\\
 
 $\checkmark$& $\times$
 & \textbf{62.64}& \textbf{81.08}& \textbf{65.66}& \textbf{84.27}& \uline{57.56}&\uline{74.10}
 & \uline{86.00}& \textbf{86.14}& 85.86& 86.75& \uline{87.04}&86.46\\
 
 $\checkmark$& $\checkmark$
 & \uline{62.52}& 80.84& 65.40& \uline{84.08}& \textbf{57.74}&\textbf{74.51}
 & \textbf{86.37}& \uline{86.02}& \textbf{86.72}& \textbf{87.14}& \textbf{88.07}&86.24\\
\bottomrule
\end{tabular}
\end{adjustbox}
\end{table*}
\begin{table*}[t]
\centering
\caption{\textbf{Ablations on the data combination during joint training.}}
\label{tab:joint}

\begin{adjustbox}{max width=\textwidth}
\begin{tabular}{cc|cccccc|ccc|ccc}

\toprule
\multirow{3}{*}{\#Row}
& \multirow{3}{*}{Data}
& \multicolumn{6}{c|}{HierText}
& \multicolumn{3}{c|}{Total-Text}
& \multicolumn{3}{c}{CTW1500} \\

\cmidrule(lr){3-8} \cmidrule(lr){9-11} \cmidrule(lr){12-14}
&
& \multicolumn{2}{c}{Word}
& \multicolumn{2}{c}{Text-line}
& \multicolumn{2}{c|}{Layout} 
& \multicolumn{3}{c|}{Word}
& \multicolumn{3}{c}{Text-line} \\

\cmidrule(lr){3-4} \cmidrule(lr){5-6} \cmidrule(lr){7-8} \cmidrule(lr){9-11} \cmidrule(lr){12-14}
&& PQ & F & PQ & F & PQ & F & F & P & R & F & P & R  \\
\midrule
 1&HierText
 & \textbf{63.50}& \textbf{81.74}& \textbf{66.96}& \textbf{85.30}& \textbf{58.35}&\textbf{74.63}& 69.02& 62.62& 76.87& 73.48& 73.01&73.96\\
 2&HierText+Total-Text+CTW1500
 & 61.96& 80.59& 64.96& 83.85& 57.18&74.16 & \uline{85.93}& \uline{85.51}& \uline{86.36}& \textbf{87.45}& \uline{87.17}&\textbf{87.72}\\
 3&Row\#2+IC13+IC15+TextSeg
 & \uline{62.52}& \uline{80.84}& \uline{65.40}& \uline{84.08}& \uline{57.74}&\uline{74.51}& \textbf{86.37}& \textbf{86.02}& \textbf{86.72}& \uline{87.14}& \textbf{88.07}&\uline{86.24}\\
\bottomrule
\end{tabular}
\end{adjustbox}
\vspace{-4mm}
\end{table*}

\noindent\textbf{Task Prompts.} 
We investigate the impact of task prompts across diverse benchmarks in \cref{tab:task prompt}. On HierText, a dataset characterized by high word density and hierarchical text structures, the integration of either task prompt yields performance gains. The simultaneous application of prompts to both the point decoder and HM-Decoder results in the best layout analysis performance. 
On Total-Text, task prompts also bring clear performance gains. The task prompt in the point decoder plays a more essential role, and combining it with the mask task prompt further improves the F-score. 
On CTW1500, using either task prompt alone slightly decreases the F-score, but employing both task prompts together still achieves the best performance.
Overall, these results underscore the effectiveness of task prompts in facilitating adaptation across diverse data distributions and varying text granularities.

\noindent\textbf{Influence of Joint Training Data.} 
Table \ref{tab:joint} presents a comparative analysis of model performance on HierText, Total-Text, and CTW1500 under varying joint training data configurations. 
Initially, when trained exclusively on HierText, the model demonstrates limited cross-dataset generalization on Total-Text and CTW1500. 
With the incorporation of Total-Text and CTW1500 in joint training, the performance on these two single-level datasets is significantly improved, with F-score improvements of 16.91\% and 13.97\%, respectively. 
This outcome underscores the flexibility and data scalability inherent in our joint training strategy.
Meanwhile, performance degradation is observed on HierText, probably attributable to data distribution shifts inherent to the relatively small-scale heterogeneous data mixing.
Subsequently, upon further integrating ICDAR2013, ICDAR2015, and TextSeg, noticeable improvements are observed on Total-Text and HierText. 
Collectively, we demonstrate that our method establishes a viable pathway for data scalability. We anticipate that expanding data volume and scenario coverage will further enhance the overall performance and robustness.

\begin{table*}[t]
\centering
\caption{\textbf{Ablations on point threshold for filtering points on word heatmap.}}
\label{tab:threshold}

\setlength{\tabcolsep}{2pt}
\begin{adjustbox}{max width=\textwidth}
\begin{tabular}{c|cccccc|ccc|ccc}

\toprule
\multirow{3}{*}{\makecell{Point\\Threshold}}
& \multicolumn{6}{c|}{HierText}
& \multicolumn{3}{c|}{Total-Text}
& \multicolumn{3}{c}{CTW1500} \\

\cmidrule(lr){2-7} \cmidrule(lr){8-10} \cmidrule(lr){11-13}
& \multicolumn{2}{c}{Word}
& \multicolumn{2}{c}{Text-line}
& \multicolumn{2}{c|}{Layout}  
& \multicolumn{3}{c|}{Word}
& \multicolumn{3}{c}{Text-line}  \\

\cmidrule(lr){2-3} \cmidrule(lr){4-5} \cmidrule(lr){6-7} \cmidrule(lr){8-10} \cmidrule(lr){11-13}
& PQ & F & PQ & F & PQ & F  
& F & P & R & F & P & R  \\
\midrule
 0.5& \uline{62.35}& \uline{80.75}& \uline{65.18}& \uline{84.06}& \textbf{57.87}& \textbf{74.88}& 84.24& 79.41& \textbf{89.70}& 86.69& 85.33&\textbf{88.10}\\
 0.6& \textbf{62.52}& \textbf{80.84}& \textbf{65.40}& \textbf{84.08}& \uline{57.74}&\uline{74.51}& \uline{85.45}& 82.53& \uline{88.57}& \uline{87.05}& 86.68&\uline{87.43}\\
 0.7& 62.17& 80.11& 64.17& 81.96& 56.36&72.43& \textbf{86.37}& \uline{86.02}& 86.72& \textbf{87.14}& \uline{88.07}&86.24\\
 0.8& 56.98& 72.73& 54.72& 69.02& 47.99&61.41& 83.73& \textbf{88.60}& 79.36& 86.41& \textbf{89.55}&83.48\\
\bottomrule
\end{tabular}
\end{adjustbox}
\vspace{-2mm}
\end{table*}
\begin{table*}[t]
\centering
\caption{\textbf{Inference latency under different point thresholds.}}
\label{tab:latency_threshold}

\setlength{\tabcolsep}{2pt}
\begin{adjustbox}{max width=\textwidth}
\begin{tabular}{c|cc|cc|cc}

\toprule
\multirow{2}{*}{\makecell{Point\\Threshold}}
& \multicolumn{2}{c|}{HierText}
& \multicolumn{2}{c|}{Total-Text}
& \multicolumn{2}{c}{CTW1500} \\

\cmidrule(lr){2-3} \cmidrule(lr){4-5} \cmidrule(lr){6-7}
& \makecell{Average\\Points}
& \makecell{Average\\Time}
& \makecell{Average\\Points}
& \makecell{Average\\Time}
& \makecell{Average\\Points}
& \makecell{Average\\Time}  \\
\midrule
 0.5& 296& 1.05s& 85& 0.50s& 109&0.50s\\
 0.6& 287& 1.02s& 82& 0.48s& 107&0.50s\\
 0.7& 263& 0.92s& 80& 0.47s& 105&0.49s\\
 0.8& 182& 0.72s& 73& 0.46s& 101&0.48s\\
\bottomrule
\end{tabular}
\end{adjustbox}
\vspace{-4mm}
\end{table*}

\noindent\textbf{Influence of Prompt Point Threshold.} 
In \cref{tab:threshold}, we present a sensitivity analysis regarding the threshold for sampling prompt points derived from the word-level heatmap.
The empirical results indicate an inherent trade-off, \myie, increasing the threshold reduces the density of sampled prompt points and consequently enhances precision at the expense of recall. 
Specifically, on HierText, a threshold of 0.6 achieves the optimal balance across various text granularities.
Conversely, Total-Text and CTW1500 benefit from a stricter threshold of 0.7 to maximize the F-score.
Consequently, our experimental protocol standardizes the point threshold at 0.6 for HierText and 0.7 for the remaining benchmarks.
We additionally study the effect of the point threshold on overall inference latency on an A800 GPU in \cref{tab:latency_threshold}. As the point threshold increases, the average number of sampled prompt points decreases, leading to a reduction in inference latency. However, the resulting improvement in efficiency is marginal in practice, therefore we set the threshold considering performance.

\noindent\textbf{Loss Weight Configuration.} 
The weighting of word, text-line, and paragraph mask losses follows Hi-SAM and the heatmap loss weight is set to 50 to balance its magnitude with other loss terms. As shown in \cref{tab:heatmap_loss_weight}, the setting of the heatmap loss weight essentially reflects a trade-off between prompt point accuracy and mask quality. When the weight is relatively small, insufficient prompt point accuracy leads to degraded performance in dense text scenarios. Conversely, when the weight is relatively large, it slightly compromises mask quality, resulting in a performance drop on both HierText and CTW1500. Therefore, we empirically set the heatmap loss weight to 50, which provides a reasonable balance between prompt point accuracy and mask quality.
\begin{table*}[t]
    \caption{\textbf{Ablations on heatmap loss weight.}}
    \label{tab:heatmap_loss_weight}
    \centering
    \resizebox{\columnwidth}{!}{
    \begin{tabular}{c|cccccc|ccc|ccc}
        \toprule
        \multirow{3}{*}{\makecell{Heatmap\\Loss Weight}}
        & \multicolumn{6}{c|}{HierText}
        & \multicolumn{3}{c|}{Total-Text}
        & \multicolumn{3}{c}{CTW1500} \\
        \cmidrule(lr){2-7} \cmidrule(lr){8-10} \cmidrule(lr){11-13}
        & \multicolumn{2}{c}{Word}
        & \multicolumn{2}{c}{Text-line}
        & \multicolumn{2}{c|}{Layout} 
        & \multicolumn{3}{c|}{Word}
        & \multicolumn{3}{c}{Text-line} \\
        \cmidrule(lr){2-3} \cmidrule(lr){4-5} \cmidrule(lr){6-7} \cmidrule(lr){8-10} \cmidrule(lr){11-13}
        & PQ & F & PQ & F & PQ & F & F & P & R & F & P & R  \\
        \midrule
        25& 61.61& 79.85& 64.90& 83.63& 56.68& 73.23& \uline{86.57}& \uline{86.07}& \textbf{87.08}& \textbf{87.37}& \uline{87.38}&\textbf{87.35}\\
        50& \textbf{62.52}& \textbf{80.84}& \textbf{65.40}& \uline{84.08}& \textbf{57.74}& \textbf{74.51}& 86.37& 86.02& \uline{86.72}& \uline{87.14}& \textbf{88.07}&\uline{86.24}\\
        100& \uline{62.06}& \uline{80.60}& \uline{65.34}& \textbf{84.23} & \uline{56.75}& \uline{73.52}& \textbf{86.91} & \textbf{87.15} & 86.68 & 86.19 & 86.59 &85.79 \\
        \bottomrule
    \end{tabular}}
    \vspace{-4mm}
\end{table*}

\section{Conclusions and Limitations}
In this work, we present ET-SAM for unified scene text detection and layout analysis, addressing the critical bottlenecks of inference latency and data utilization that may hinder the practical application of SAM-based frameworks. 
Instead of relying on dense, random foreground point sampling, ET-SAM leverages a tailored lightweight decoder to derive sparse, high-confidence points from word-level heatmaps, successfully eliminating excessive computational overhead.
Furthermore, we effectively harmonize heterogeneous annotations in diverse datasets via designing a joint training strategy and introducing task prompt tokens.
Extensive experiments demonstrate that, compared to Hi-SAM, ET-SAM achieves approximately \textbf{3$\times$} inference acceleration with competitive performance on HierText and realizes a substantial 11.0\% improvement in average F-score across Total-Text, CTW1500, and ICDAR2015.

However, ET-SAM is primarily designed for text detection and cannot be directly applied to high-level text understanding tasks, as its learned visual features are not aligned with vision-language models, limiting its end-to-end multimodal application. Future work could explore incorporating semantic modeling and cross-modal alignment to address this limitation.

\section*{Acknowledgements}
This work was supported in part by the New Generation Artificial Intelligence-National Science and Technology Major Project under Grant 2025ZD0123602, in part by the National Natural Science Foundation of China under Grants 62225113, U23B2048 and 625B2132, in part by the Innovative Research Group Project of Hubei Province under Grant 2024AFA017, in part by the Science and Technology Major Project of Hubei Province under Grants 2024BAB046 and 2025BCB026, and in part by the New Cornerstone Science Foundation through the XPLORER PRIZE. This work was also supported by WHU-Kingsoft Joint Lab. The numerical calculations in this paper have been done on the supercomputing system in the Supercomputing Center of Wuhan University.

\clearpage
%
%
\bibliographystyle{splncs04}
\bibliography{main}
\clearpage

\appendix
\section{Center Point Heatmap}
We have already introduced a method for generating the target heatmap based on the center lines of word instances. In addition, we investigate an alternative strategy that generates the target heatmap from word center points using an anisotropic Gaussian kernel.

Specifically, we first compute the minimum bounding rectangle for each word contour according to the word annotations. Given the bounding rectangle, we obtain the center coordinates $P(P_x, P_y)$, the rotation angle $\theta$, as well as the height $h$ and width $w$. Based on these parameters, the heatmap for each word center point is generated using an anisotropic Gaussian kernel:
\begin{equation}
\begin{aligned}
H_{xy} = \exp \Bigg( - \Bigg(
& \frac{\left((x - p_x)\cos\theta + (y - p_y)\sin\theta\right)^2}{2\sigma_h^2} \\
& + \frac{\left((x - p_x)\sin\theta - (y - p_y)\cos\theta\right)^2}{2\sigma_w^2}
\Bigg) \Bigg),
\end{aligned}
\end{equation}
where $(p_x,p_y)=(\frac{P_x}{4},\frac{P_y}{4})$ denote the downsampled coordinates of point $P$ on the heatmap grid, $(x,y)$ represents an arbitrary spatial coordinate on the heatmap, and the standard deviations $\sigma_h$ and $\sigma_w$ are proportional to the height $h$ and width $w$ of the word bounding rectangle, respectively. 
Finally, the heatmaps of all words are aggregated to form the center point heatmap $H$, where the maximum value is taken at each grid location.

\begin{table*}[b]
\centering
\caption{\textbf{Comparison of Target Heatmap on HierText.}`$\dagger$' and `$\ddagger$' indicate training model on HierText using the center line heatmap and center point heatmap, respectively, to generate prompt points.}
\label{tab:point heatmap}

\setlength{\tabcolsep}{6pt}
\resizebox{\linewidth}{!}{\begin{tabular}{ll|cc|cc|cc|c|c}
\toprule
\multirow{2}{*}{Dataset}
 & \multirow{2}{*}{Method}
 & \multicolumn{2}{c|}{Word} 
 & \multicolumn{2}{c|}{Text-line} 
 & \multicolumn{2}{c|}{Layout} 
 & \multirow{2}{*}{\makecell{Average\\Points}}
 & \multirow{2}{*}{\makecell{Average\\Time}}\\

\cmidrule(lr){3-4} \cmidrule(lr){5-6} \cmidrule(lr){7-8}
& & PQ & F & PQ & F & PQ & F & & \\

\midrule
\multirow{3}{*}{\makecell{Val. Set}} & Hi-SAM-L 
&  62.60&  \uline{80.99}&  \uline{65.62}&  \uline{84.04}&  58.83&  75.84&  1500&  4.4s\\

&Hi-SAM-L$^{\dagger}$
& \uline{62.77}& 80.65& \textbf{66.40}& \textbf{84.54}& \textbf{59.81}& \textbf{76.36}&  \uline{269}&  \uline{1.2s}\\

&Hi-SAM-L$^{\ddagger}$
& \textbf{62.89}& \textbf{80.99}& 65.59& 83.89& \uline{59.47}& \uline{76.16}&  \textbf{155}&  \textbf{1.0s}\\

\midrule
\multirow{3}{*}{\makecell{Test Set}} & Hi-SAM-L 
&  63.10&  \uline{81.83}&  66.17&  84.85&  57.61&  74.49&  1500&  4.4s\\

&Hi-SAM-L$^{\dagger}$
& \uline{63.50}& 81.74& \textbf{66.96}& \textbf{85.30}& \uline{58.35}& \uline{74.63}&  \uline{268}&  \uline{1.2s}\\

&Hi-SAM-L$^{\ddagger}$
& \textbf{63.67}& \textbf{82.13}& \uline{66.36}& \uline{84.87}& \textbf{58.45}& \textbf{75.08}&  \textbf{153}&  \textbf{1.0s}\\
\bottomrule
\end{tabular}}
\end{table*}

\begin{figure*}[t!]
    \centering
    \includegraphics[width=0.8\textwidth]{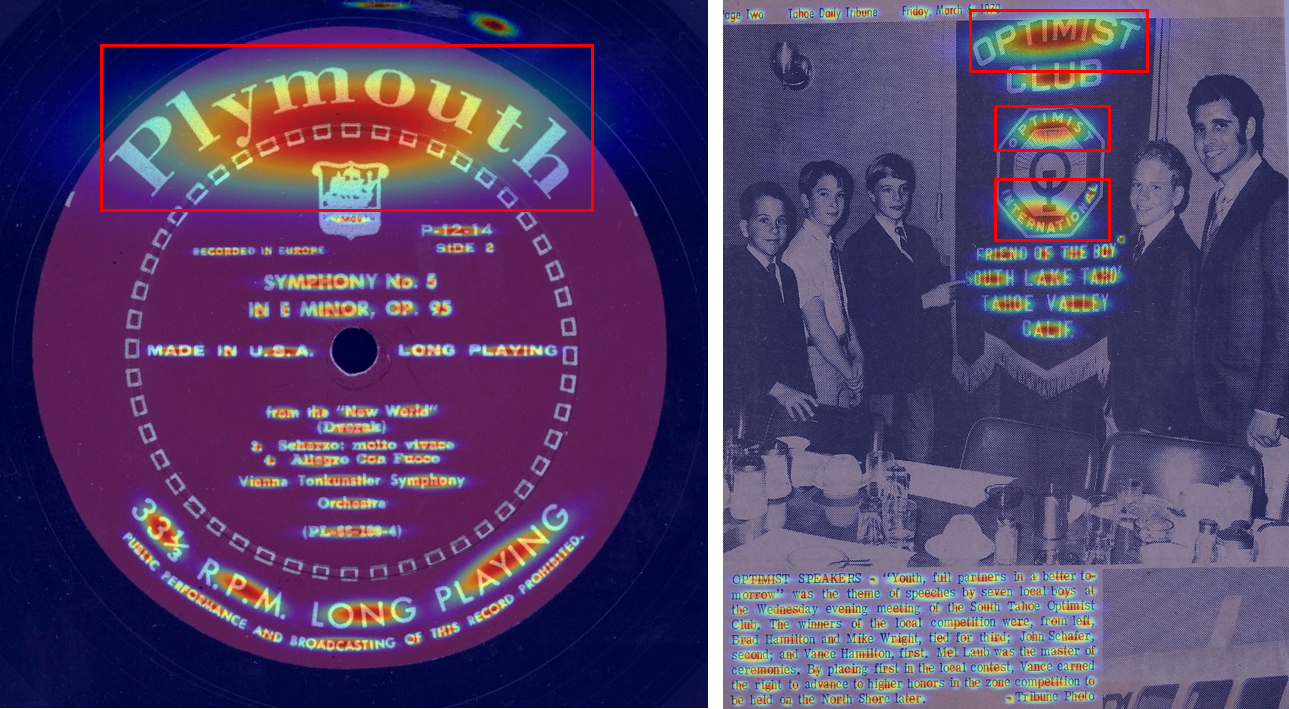}
    \caption{\textbf{Visualizations of the Center Point Heatmap.} The heatmap centers of the curved text highlighted in the red boxes are offset from the word center regions}
    \label{fig:visualize_point_heatmap}
    \vspace{-4mm}
\end{figure*}

We compare the performance and efficiency of the center point heatmap and center line heatmap methods on HierText, with results presented in \cref{tab:point heatmap}. 
Overall, the point–based approach achieves better performance at the word level, while the line–based method performs better at the text-line level. 
The former generates roughly 110 fewer prompt points than the latter and achieves an inference speedup of about 0.2s.
However, the center point heatmap is not suitable for curved text. 
As shown in \cref{fig:visualize_point_heatmap}, the heatmap centers of curved text in the red boxes do not align with the word center regions. 
Since curved text is relatively rare in HierText, the performance difference between the two methods shown in \cref{tab:point heatmap} is minimal. 
In contrast, Total-Text and CTW1500 contain a larger proportion of curved text, and continuing to use the center point heatmap for joint training or fine-tuning would likely degrade performance. 
Therefore, we adopt the center line heatmap as the target heatmap.

\section{More Comparisons on Scene Text Detection}
Considering that the scale and component of our training data are different from some SOTA methods, we train DPText-DETR using the same datasets as ET-SAM to ensure a fair comparison. 
Concretely, for Total-Text and ICDAR15, we pre-train the model using word-level annotations from HierText, together with Total-Text, TextSeg, ICDAR13, and ICDAR15. For CTW1500, we pre-train the model using text-line annotations from HierText, along with Total-Text, TextSeg, ICDAR13, ICDAR15, and CTW1500. 
The two pre-training stages adopt the same settings, with a batch size of 8 and a total of 350k iterations. The learning rate is set to $1e^{-4}$ and reduced to $1e^{-5}$ at 280k.
Finally, the pre-trained models are fine-tuned on each target dataset individually.
For Total-Text, the model is fine-tuned for 20k iterations with an initial learning rate of $5e^{-5}$, which decays to $5e^{-6}$ at 16k. 
For CTW1500, we fine-tune the model for 13k iterations with a learning rate of $2e^{-5}$. 
For ICDAR15, the model is fine-tuned for 20k iterations with an initial learning rate of $1e^{-5}$, which is reduced to $1e^{-6}$ at 16k.

We evaluate the pre-trained and fine-tuned models from the above experiments and compare them with ET-SAM, with the results presented in \cref{tab:fair comparison}. 
With the same pre-training datasets, ET-SAM (Joint) significantly outperforms DPText-DETR* (Pre-train) on Total-Text and CTW1500, demonstrating the superior data efficiency of ET-SAM. 
After fine-tuning, it achieves 2.3\% and 1.7\% F-score improvements on Total-Text and CTW1500, respectively, while showing slightly lower performance only on ICDAR15.
\begin{table*}[t]
\centering
\caption{\textbf{Comparison of methods trained on the same dataset.} `*' indicates that the model is pre-trained and fine-tuned using the same datasets as ET-SAM.}
\label{tab:fair comparison}

\setlength{\tabcolsep}{6pt}
\vspace{-2mm}

\resizebox{\linewidth}{!}{
\begin{tabular}{l|ccc|ccc|ccc}
\toprule
\multirow{2}{*}{Method}
 & \multicolumn{3}{c|}{Total-Text} 
 & \multicolumn{3}{c|}{CTW1500} 
 & \multicolumn{3}{c}{ICDAR15} \\

\cmidrule(lr){2-4} \cmidrule(lr){5-7} \cmidrule(lr){8-10}
& F & P & R
& F & P & R
& F & P & R \\

\midrule

DPText-DETR
&\textbf{89.0}	&\textbf{91.8}	&86.4	&\textbf{88.8}	&\textbf{91.7}	&86.2	&-	&-	&-\\

DPText-DETR* (Pre-train)
&77.5&73.7&81.7&67.2&59.2&77.7&\textbf{83.7}&\uline{94.7}&75.1\\

DPText-DETR* (FT)
&84.7&84.3&85.1&86.2&84.7&\uline{87.7}&\uline{83.7}&\textbf{94.7}&75.0\\

ET-SAM (Joint) 
& 86.4& 86.0& \uline{86.7}& 87.1& \uline{88.1}& 86.2&  83.5&  85.0& \uline{82.1}\\

ET-SAM (FT) 
&\uline{87.0}&\uline{86.4}&\textbf{87.5}&\uline{87.9}&87.3&\textbf{88.4}&83.2&80.6&\textbf{85.9}\\
\bottomrule
\end{tabular}}
\end{table*}

\section{Broader Efficiency Analysis}
We provide a detailed breakdown analysis in \cref{fig:introduction}. Reducing excessive point prompts highly accelerates the hierarchical segmentation process (from 3.73s to 0.94s, on V100) and subsequent post-processing operation (from 0.33s to 0.18s), achieving an overall 3$\times$ speedup. Considering the same model, fewer prompt counts lead to higher efficiency. However, simply reducing prompt points degrades the performance, as shown in the first two lines from \cref{tab:point decoder} whereas introducing our point decoder improves efficiency without sacrificing accuracy. 

\begin{table*}[t]
\caption{\textbf{Comparison of inference latency with DETR-based methods.}}
\label{tab:latency_comparison}
\centering
\setlength{\tabcolsep}{2pt}
\vspace{-2mm}
\begin{tabular}{l|ccc}
\toprule
Major Process & ET-SAM & DPText & SRFormer \\
\midrule
Image Encoding & 0.291s & 0.015s & 0.015s \\
\midrule
\makecell[l]{Mask/Box\\Decoding} & \makecell{0.580s\\(HM-Decoder)} & \makecell{0.038s\\(Trans.)} & \makecell{0.114s\\(Trans.)} \\
\bottomrule
\end{tabular}
\vspace{-4mm}
\end{table*}
Moreover, as shown in \cref{tab:latency_comparison}, compared to DETR-based detectors (measured on HierText using an A800 GPU), ET-SAM consumes higher image encoding overhead due to the heavy backbone while batch-wise mask decoding emerges as the dominant latency bottleneck compared to the Transformer encoder and decoder in the DETR framework.

\end{document}